\documentclass[11pt,a4paper]{article}
\usepackage[hyperref]{emnlp2020}
\usepackage[american]{babel}
\usepackage[T1]{fontenc}
\usepackage{times}
\usepackage{url}
\usepackage{microtype}

\aclfinalcopy 

\usepackage{amsmath}
\DeclareMathOperator*{\argmax}{arg\,max}

\newcommand{\Ni}{(1)~}
\newcommand{\Nii}{(2)~}
\newcommand{\Niii}{(3)~}

\newcommand{\Na}{(a)~}
\newcommand{\Nb}{(b)~}
\newcommand{\Nc}{(c)~}

\RequirePackage{color}
\definecolor{darkgray}{gray}{0.40}
\definecolor{mediumgray}{gray}{0.60}
\definecolor{lightgray}{gray}{0.95}
\definecolor{ultralightgray}{gray}{0.98}

\usepackage{graphicx}
\DeclareGraphicsExtensions{.pdf,.ai,.jpg,.png}
\setkeys{Gin}{pagebox=artbox}
\graphicspath{{emnlp20-bias-detection-paper-figures/}}

\newcommand{\bsfigure}[3][scale=1.0]{%
  \begin{figure}[tb]
    \centering
    \includegraphics[#1]{#2}
    \caption{#3}\label{#2}
  \end{figure}}

\newcommand{\hwfigure}[3][t!]{%
  \begin{figure*}[#1]
    \centering
    \includegraphics[scale=1.0]{#2}
    \caption{#3}\label{#2}%
  \end{figure*}}

\usepackage{booktabs}
\RequirePackage{type1cm}
\RequirePackage{color}
\RequirePackage{soul}
\setstcolor{blue}

\newsavebox\bscombox
\newcommand{\bscom}[3][]{%
  \sbox{\bscombox}{\fontsize{8}{9}\selectfont#1#2#3}
  \noindent
  \st{#2}{\selectfont
    \color{blue}#3\ifx\\#1\\\else{\fontsize{8}{9}\selectfont\color{violet}[#1]}\fi
    }
  }

\begin{document}

\title{Detecting Media Bias in News Articles using Gaussian Bias Distributions}

\author{
Wei-Fan Chen \\
Paderborn University \\
Department of Computer Science \\
{\tt cwf@mail.upb.de} \\\And
Khalid Al-Khatib \\
Bauhaus-Universit\"at Weimar \\
Faculty of Media, Webis Group \\
{\tt khalid.alkhatib@uni-weimar.de } \\\AND
Benno Stein\\
Bauhaus-Universit\"at Weimar \\
Faculty of Media, Webis Group \\
{\tt benno.stein@uni-weimar.de} \\\And
Henning Wachsmuth \\
Paderborn University \\
Department of Computer Science \\
{\tt henningw@upb.de}
}

\date{}

\maketitle

\begin{abstract}
Media plays an important role in shaping public opinion. Biased media can influence people in undesirable directions and hence should be unmasked as such. We observe that feature-based and neural text classification approaches which rely only on the distribution of low-level lexical information fail to detect media bias. This weakness becomes most noticeable for articles on new events, where words appear in new contexts and hence their ``bias predictiveness'' is unclear. In this paper, we therefore study how second-order information about biased statements in an article helps to improve detection effectiveness. In particular, we utilize the probability distributions of the frequency, positions, and sequential order of lexical and informational sentence-level bias in a Gaussian Mixture Model. On an existing media bias dataset, we find that the frequency and positions of biased statements strongly impact article-level bias, whereas their exact sequential order is secondary. Using a standard model for sentence-level bias detection, we provide empirical evidence that article-level bias detectors that use second-order information clearly outperform those without. 
\end{abstract}

\section{Introduction}

Media bias is discussed and analyzed in journalism research \cite{groseclose:2005,dellavigna:2007,iyengar:2009} and NLP research \cite{gerrish:2011,iyyer:2014,chen:2018}. According to the study of \newcite{groseclose:2005}, bias \emph{``has nothing to do with the honesty or accuracy''}, but it means \emph{``taste or preference''}. In fact, journalists may
\Ni
report facts only in favor of one particular political side and thus
\Nii
conclude with their own opinion.
As an example, the following sentences from {\em allsides.com} reporting on the event ``Trump asks if disinfectant, sunlight can treat coronavirus'' demonstrate media bias on the sentence level:

\smallskip
{\em The activists falsely claimed that Trump ``urged Americans to inject themselves with disinfectant'' and ``told people to drink bleach.''}

\noindent
--- The Daily Wire, right-oriented

\smallskip
{\em Lysol maker issues warning against injections of disinfectant after Trump comments}

\noindent
--- The Hill, center-oriented

\smallskip
{\em ``This notion of injecting or ingesting any type of cleansing product into the body is irresponsible and it's dangerous,'' said Gupta.}

\noindent
---  NBC News, left-oriented

\medskip
From an NLP perspective, bias in the example sentences could be detected by capturing sentiment words, such as ``falsely'' or ``irresponsible''. Without the background knowledge of the political side of Trump or the event itself, however, predicting which side these sentences are slanted to is difficult. 

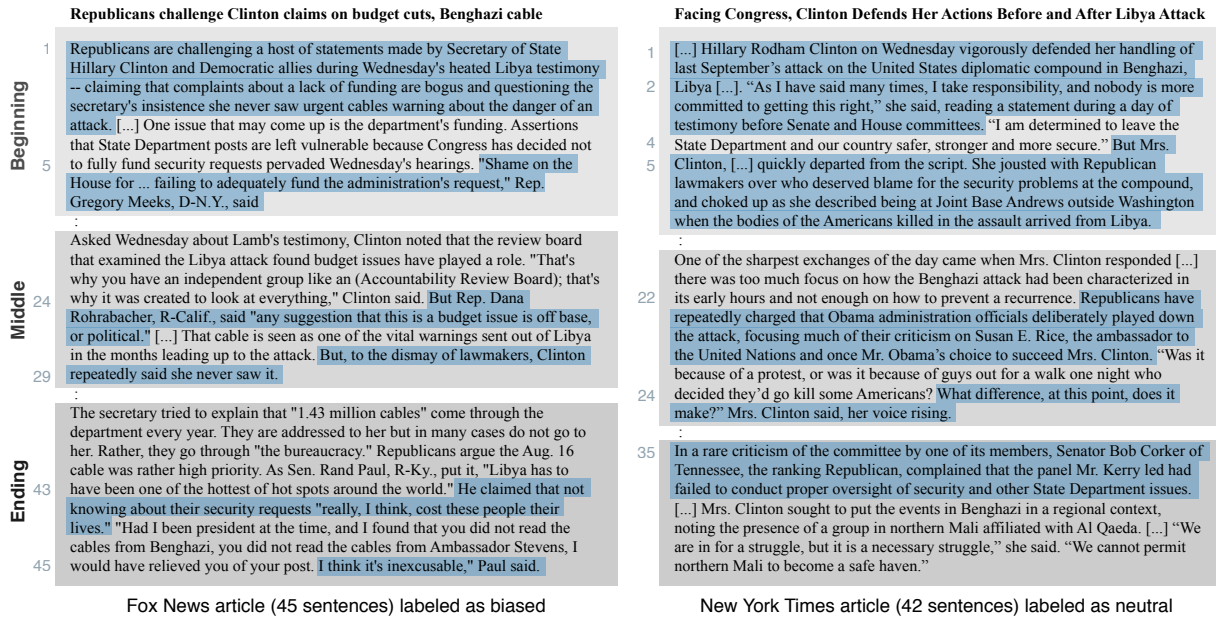
\hwfigure{example-articles}{Excerpts of a biased article (left) and a neutral article (right) from the used dataset. All sentences labeled as having lexical or informational bias are highlighted; their position can be read from the numbers next to them.}

Bias detection even becomes harder at the article level. For illustration, Figure~\ref{example-articles} shows two articles and their sentence-level bias from the used dataset. It becomes clear that the actual words in the biased sentences are not always indicative to distinguish biased from neutral articles, nor is the count of the biased sentences: Bias assessments on sentence level do not ``add up''. In this regard, the {\em position} of biased sentences appears to be a better feature.

The existing approaches to bias detection are transferred from other, less intricate text classification tasks. They largely model \emph{low-level lexical information}, either explicitly, e.g.\ by using bag-of-words \cite{gerrish:2011}, or implicitly via neural networks \cite{gangula:2019}. Such approaches tend to fail at the article level, particularly for articles on events not covered in the training data. The reason is that bias clues are subtle and rare in articles, especially event-{\em in}dependent clues. Altogether, modeling low-level information at the article level is insufficient to detect article-level bias, as we will later stress in experiments.

We study article-level bias detection both with and without allowing to learn event-specific information. The latter scenario is more challenging, but it is closer to the real world, because we cannot expect that the information in future articles always relates to past events. Inspired by ideas from modeling local and global polarities in sentiment analysis \cite{wachsmuth:2015b}, we hypothesize that using {\em second-order bias information} in terms of lexical and informational bias at the sentence level is key to detecting article-level bias. To the best of our knowledge, no bias detection approach so far uses such information. We investigate this hypothesis in light of three research questions:

\begin{enumerate}
\setlength{\itemsep}{0pt}
\item[Q1.] 
How effective are standard classification approaches in article-level bias detection, with and without exploiting event information?
\item[Q2.] 
How does sentence-level bias impact article-level bias in general? 
\item[Q3.]
To what extent can sentence-level bias detection be utilized for article-level bias detection?
\end{enumerate}

To study Q1--Q3, we employ the BASIL dataset, which includes manually annotated bias labels at article level as well as lexical and informational bias labels at sentence level \cite{fan:2019}. While the dataset contains only 300 articles, it provides the best basis for understanding the interaction of bias at both levels available so far.

For Q1, we evaluate an $n$-gram-based SVM and a BERT-based neural network in article-level bias detection. To assess the impact of event-related information, we split the dataset in two ways, once with event overlap in the training set and test set, and once without. As expected, we observe that the effectiveness of both approaches is generally low, especially when event information cannot be exploited. The results indicate that the concept of sentence-level bias is too subtle and rare to be utilized by these approaches.

For Q2, we study multiple types of correlations between sentence-level and article-level bias on the ground-truth annotations, covering
\Na
the frequency of biased sentences,
\Nb
their position in an article, and
\Nc
their sequential order.
For each type, we model the bias distribution in a new way through a Gaussian Mixture Model (GMM), in order to then exploit it as features of an SVM (for frequency), Na\"{i}ve Bayes (for positions), and a first-order Markov model (for sequential order). The results show strong correlations between the two levels for frequency and position information, whereas sequential order seems less correlated.

For Q3, finally, we propose a new approach applicable in realistic settings. In particular, we retrain the bias detectors from the Q1 experiments on the sentence level and then exploit the GMM as above to predict to article level bias. In our evaluation, the approach significantly outperforms  the article-level approaches analyzed for Q1. Countering intuition, it even achieves higher effectiveness than what we observed on the ground truth for Q2. We explain this result by the fact that the sentence-level detector creates more deterministic sentence bias features, allowing our approach to learn from them in a more robust way.

Altogether, the contribution of this paper is three-fold: 
\Ni
We provide evidence that standard approaches fail in detecting article-level bias.
\Nii
We develop a new approach utilizing second-order bias information, i.e., sentence-level bias.
\Niii
We show that second-order bias information is an effective means to build better article-level bias classifiers.

\section{Related Work}
\label{sec:relatedwork}

Media bias detection has been studied with computers since the work of \citet{lin:2006}. As of then, media bias has been investigated in slight variations under different names, including {\em perspective} \cite{lin:2006}, {\em ideology} \cite{iyyer:2014}, {\em truthfulness} \cite{rashkin:2017}, and {\em hyperpartisanship} \cite{kiesel:2019}. To detect bias, early approaches relied on low-level lexical information. For example, \citet{greene:2009} used \emph{kill} verbs and \emph{domain-relevant} verbs to detect articles being pro Israeli or Palestinian perspectives. \citet{recasens:2013} relied on linguistic cues, such as factoid verbs and implicatives, in order to assess whether a Wikipedia sentence conveys a neutral point of view or not. Besides the NLP community, also researchers in journalism have approached the measurement of media bias. E.g.,  \citet{gentzkow:2010} used the preferences of phrases at each side (such as ``war on terror'' for Republican but ``war in Iraq'' for Democratic). \citet{groseclose:2005} used the counts of think-tank citations to estimate the bias.

With the rise of deep learning, NLP researchers have also used neural-based approaches for bias detection. \citet{iyyer:2014} used RNNs to aggregate the polarity of each word to predict sentence-level bias based on parse trees. \citet{gangula:2019} made use of headline attention to classify article bias. \citet{li:2019} encoded social information in their Graph-CNN. While deep learning is believed to capture deeper relations among its inputs, we show that extending a neural network from sentence-level to article-level bias detection does not ``just work''.

One point of variation in media bias detection is the level of text being analyzed, which varies from tokens \cite{fan:2019} and sentences \cite{bhatia:2018} to articles \cite{kulkarni:2018}, sources \cite{baly:2019}, and users \cite{preoctiuc:2017}. While the effectiveness of machine learning models on different levels helps understanding how media bias becomes manifest at different levels, \citet{lin:2006} are to our knowledge the only to discuss the difference between sentence-level and article-level bias detection. 

Source-level and user-level bias can be seen as directly emerging from summing up bias in the associated texts. For example, \citet{baly:2019} averaged the feature vectors of articles as the feature vectors of a source. The relation between sentence-level and article-level bias remains unstudied so far. The goal of this paper is not to discuss the difference between these levels. Rather, we examine how to aggregate the sentence-level bias to generate second-order features, and then use these features to predict article-level bias.

The use of low-level information to generate second-order features was studied in the context of product reviews by modeling patterns in the reviews' sentiment flow \cite{wachsmuth:2015b}, by tuning neural network to capture important sentences \cite{xu:2016}, and by routing in aggregating sentence embeddings into document embedding \cite{gong:2018}. In particular, our usage of low-level information is inspired by \citet{wachsmuth:2015b}, where we hypothesize that such flows exist in media bias as well. However, we do not limit our approach to entire sequences of sentence-level information, but we also consider frequency, position, or only two to three continuous sentences.

\section{Standard Bias Detection Approaches}
\label{sec:std}

Standard approaches for bias detection, on both article and sentence level, mainly exploit the low-level lexical features to classify the texts as biased or not, neglecting bias-specific features. The two main low-level lexical feature types that are employed in such approach ares:
\Ni
\emph{n}-gram features, where \emph{n} is typically one to three (i.e., unigram, bigram, or trigram), and
\Nii
word embeddings, especially within pre-trained language models (i.e., transformers) such as BERT.

We propose two classification settings to answer research question~Q1, which addresses the importance of event information: In the first setting, called {\em event overlapping}, we form the training and test sets by randomly assigning examples to them, more specifically, without looking at event information. The setting allows texts of the same event to occur in both the training and the test set. The second setting is called {\em event non-overlapping} since the texts to be classified are first categorized according to the main event that they address. During the splitting in training set and test set we then ensure for each event that all its related texts are in exactly one of these sets.

The difference in the effectiveness of the standard approaches on the two settings indicates whether and to what extent standard bias detection approaches rely on event information.

\section{Second-Order Bias Information}
\label{sec:approach}

For research question Q2, we study the correlation between sentence-level and article-level bias. Specifically, we examine whether article-level bias correlates with
\Na
the frequency of biased sentences,
\Nb
their position in an article, and
\Nc
their sequential order.
For each correlation, we extract features and then train a respective machine learning model. The code is available at \url{https://github.com/webis-de/EMNLP-20}.

\subsection{Bias Frequency}

A straightforward way of leveraging sentence-level bias information is counting. Let an article with sentence-level bias labels $\{b_1, b_2, ..., b_n\}$ be given, where $n$ is the number of sentences in the article and $b_i$ the label of the $i$-th sentence. Assuming that $b_i$ is binary with $b_i = 1$ being bias, the {\em absolute bias frequency}, $f_{abs}$, is defined as:
\begin{equation}
f_{abs} = \sum_{i=1}^{n} b_i
\end{equation}

Accordingly, the {\em relative bias frequency}, $f_{rel}$, is defined based on the length of the article as: 
\begin{equation}
f_{rel} = \frac{\sum_{i=1}^{n} b_i}{n}
\end{equation}

\subsection{Bias Position}

We consider the positions of biased sentences as second-order features. Given a target number of positions, $k$, we first normalize the sentence-level bias annotations $\{b_1, b_2, ..., b_n\}$ into $\{\bar{b}_1, \bar{b}_2, ..., \bar{b}_k\}$, with $\bar{b}_i \in [0, 1]$. The higher~$\bar{b}_i$, the more likely position $i$ is biased. In detail, we first normalize $\{b_1, b_2, ..., b_n\}$ to $\{b'_1, b'_2, ..., b'_m\}$ by linear interpolation, where $m$ (here set to 100) is larger than the largest $n$ (and also larger than $k$). After the interpolation, $b'_i$ is in the range of $[0, 1]$. Secondly, we ``sample'' from the $b_i'$ to make the final sentence-level bias having length $k$. There are three ``sampling'' methods we explore:
\Ni
average (take the average of the datapoints,
\Nii
maximum (take the maximum value in the range, and
\Niii
last (take the last datapoints).
We treat this as a hyperparameter and find the best one by the validation set. We use this two-step normalization (upsampling and then downsampling) to avoid the instability during sampling when $n/k$ is not an integer.

Our goal is to predict the most likely article-level bias label, $a^*$, given the sentence-level bias. Formally, assuming that an article can be seen as a combination of its sentences, we have
\begin{equation}
a^* = \argmax_a p(a\mid\bar{b}_1, \bar{b}_2, ..., \bar{b}_k),
\end{equation}

\noindent
where $a$ is any possible bias label (0 for neutral and 1 for bias), and $p(a\mid\cdot)$ is the conditional probability of $a$, given a sentence-level bias sequence. According to Bayes' rule and given that $p(\bar{b}_1, \bar{b}_2, ..., \bar{b}_k)$ is irrelevant to the arg\,max, we can rewrite it as:
\begin{equation}
\label{equation:bayes}
a^* = \argmax_a p(\bar{b}_1, \bar{b}_2, ..., \bar{b}_k\mid a) \cdot p(a)
\end{equation}

Assuming that each $\bar{b}_i$ is independent from other positions, we further simplify this as
\begin{equation}
\label{equation:nb}
a^* = \argmax_a \prod_{i=1}^{k}p(\bar{b}_i\mid a) \cdot p(a),
\end{equation}

\noindent
which is a Na\"{i}ve Bayes classifier, and each $p(\bar{b}_i\mid a)$ is the bias position feature we are interested in.

In the remainder, we simplify the notation $p(\bar{b}_i\mid a)$ to $p(\bar{b}\mid a)$. Estimating $p(\bar{b}\mid a)$ in each position for each $a$ is difficult, since $\bar{b} \in [0,1]$ and we cannot observe enough data points in that range on realistic text corpora. Instead, we therefore estimate $p(a\mid\bar{b}) / p(a)$, where $p(a)$ can be properly estimated by the distribution of the labels, and $p(a\mid\bar{b})$ can be estimated well using a Gaussian Mixture Model.

\paragraph{Gaussian Mixture Model}

Given a set of $m$ articles along with their bias labels, $\{a_1, a_2, ... , a_m\}$, we first retrieve the interpolated bias value in each position $b_{i,j}$ where $i$ is the index of the position and $j$ is the index of the article. $b_{i,j}, 1 \leq j \leq m$ can be seen as a distribution of the bias strength in one position $i$. For example, the distribution in Figure~\ref{gmm} shows the bias in the second position if we normalize the articles into 10 positions.

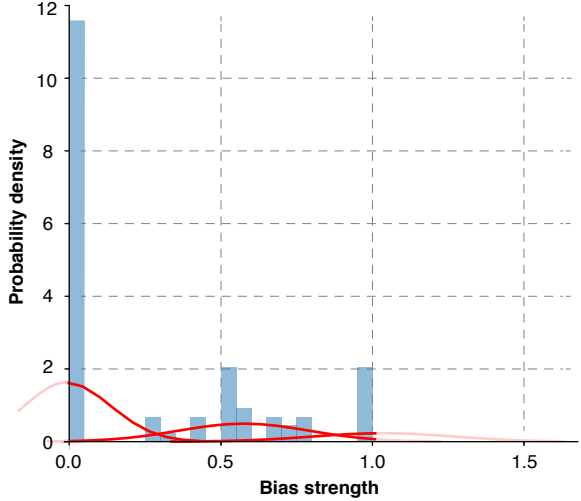
\bsfigure{gmm}{Bias strength in one position and the fitted Gaussian mixtures of it. The bias strength is the value of $\bar{b}_i$. Note that the y-axis is the probability density, i.e., the sum of all area in bins or sum of all area under Gaussian mixtures is one.}

To model the distribution, we employ a Gaussian mixture model (GMM)~\cite{reynolds:2009}. The assumption behind GMMs is that a distribution can be seen as a combination of Gaussian distributions, where each distribution is represented by its mean $\mu$, its variance $\sigma^2$, and a weight $w$, the sum of all weights being 1. Modeling a GMM is unsupervised; we only need to set the number of mixtures we would like to have.

After applying GMM on $b_{i,j}, 1 \leq j \leq m$, the distribution of a bias position $i$ is represented by a set of Gaussian mixtures, $\mathcal{N}_l(\mu_l, \sigma_l^2, w_l)$, where $l$ is the index of mixtures. For each mixture, we can then learn its bias distribution by:
\begin{equation}
p(a = 1\mid\mathcal{N}_l) = \frac{occur(\bar{b}_{i,j} \in \mathcal{N}_l, a_j = 1)}{occur(b_{i,j} \in \mathcal{N}_l)}
\end{equation}

To avoid zero probability in some mixtures, we also apply add-one smoothing. Then, the bias probability $p(\bar{b}\mid a=1)$ in one position is:
\begin{equation}
p(\bar{b}\mid a=1) \propto \frac{p(a=1|\bar{b})}{p(a = 1)} \sim \frac{p(a=1|\mathcal{N}_{\bar{b}})}{p(a = 1)}, 
\end{equation}

\noindent
where $\mathcal{N}_{\bar{b}}$ is the mixture most likely generating $\bar{b}$.

\subsection{Bias Sequence}

The Na\"{i}ve Bayes classifier in Equation~\ref{equation:nb} assumes that each position is independent from other positions. We can also consider a position to depend on the previous positions. For example, under the assumption that each position depends on the one before, we can rewrite Equation~\ref{equation:nb} as:
\begin{equation}
\label{equation:markov}
a = \argmax_a \prod_{i=1}^{k}p(\bar{b}_i\mid\bar{b}_{i-1}, a) \cdot p(a)
\end{equation}

Then, we can further rewrite $p(\bar{b}_i\mid\bar{b}_{i-1}, a)$ as:
\begin{equation}
p(\bar{b}_i\mid\bar{b}_{i-1}, a) = \frac{p(a\mid\bar{b}_{i}, \bar{b}_{i-1})}{p(\bar{b}_{i-1}\mid a) \cdot p(a)}
\end{equation}

In this equation, $p(\bar{b}_{i-1}\mid a)$ can be approached by the GMM as described, and the numerator of the equation can be seen as the transition probability in a Markov process. In particular, after finding the mixtures most likely generating $\bar{b}_{i}$, and $\bar{b}_{i-1}$, we estimate the transition probability $p(a|\bar{b}_{i}, \bar{b}_{i-1})$ as:
\begin{equation}
p(a\mid\bar{b}_{i}, \bar{b}_{i-1}) \sim p(a\mid\mathcal{N}_i,\mathcal{N}_{i-1}),
\end{equation}

\noindent
where $\mathcal{N}_i$ and $\mathcal{N}_{i-1}$ are the mixtures most likely generating $\bar{b}_{i}$ and $\bar{b}_{i-1}$ respectively. Again, we apply add-one smoothing when estimating the transition probabilities. 

The previous equations can be easily extended to the case that each position is dependent on more than one position. However, longer dependencies imply fewer observations of each possible transition. As a result, we only test the first and the second-order Markov process below (i.e., dependence on the previous one or two positions).

\section{Experiments}
\label{sec:exp} 

This section presents the experiments that we designed to study research questions Q1--Q3 based on the media bias dataset BASIL.

\subsection{Dataset}

To test the hypothesis that sentence-level bias is an important feature for article-level bias detection, we need data that is annotated for both bias levels. Recently, \citet{fan:2019} released a dataset on media bias, \emph{Bias Annotation Spans on the Informational Level (BASIL)}. The dataset contains 300 news articles on 100 events, three each per event. These three articles were taken from Fox News, New York Times, and Huffington Post, which have been selected as a representative of right-oriented, neutral, and left-oriented portals respectively.

On the article level, the dataset comes with manually annotated media bias labels (right, center, or left). While we noticed that more Fox news articles are right (50) than Huffingtion post articles (10), the labels do not only rely on the source of the articles. Since we target bias in general rather than a specific orientation, we merged right and left to the label {\em bias}, and see center as {\em neutral}. Because both bias and unbiased articles include all three portals, we can be confident that the task is not detecting the source, but detecting the bias.

On the sentence level, each sentence has been manually labeled as having \emph{lexical bias}, \emph{informational bias}, or \emph{none}. According to \citet{fan:2019}, lexical bias refers to ``how things are said'', i.e., the author used polarized or otherwise sentimental words showing bias. On the other hand, sentences with informational bias ``convey information tangential or speculative''. In our experiments, we considers both settings where we separate the two types of bias and settings where we merge them.

\subsection{Experiment Settings}

In light of our three research questions, we consider the following experiments: 

\begin{table}[t!]

\small
\centering
\setlength{\tabcolsep}{4pt}

\begin{tabular*}{\linewidth}{l rr rr rr}
\toprule
& \multicolumn{2}{c}{\bf Training} & \multicolumn{2}{c}{\bf Validation} & \multicolumn{2}{c}{\bf Test} \\ 
\cmidrule(l@{2pt}r@{2pt}){2-3} \cmidrule(l@{2pt}r@{2pt}){4-5} \cmidrule(l@{2pt}r@{2pt}){6-7}
& Neutral & Bias & Neutral & Bias & Neutral & Bias \\
\midrule
w/ Event & 85 & 95 & 26  & 34 & 33 & 27 \\
w/o Event & 84 & 96 & 31  & 29 & 29 & 31 \\
\bottomrule
\end{tabular*}

\caption{Bias distribution of articles in the two experiment settings for research question~Q1: \emph{w/ event} indicates that there is event overlap in the training, validation, and test set (random split), while \emph{w/o event} refers to an event-controlled split.}
\label{table-art-level}

\end{table}

\paragraph{Q1.} 

To study Q1, we compare two experiment settings of article-level bias detection:
\Ni
with event information being available, and 
\Nii
with event information not being available.
In both settings, the size of the training set (180 articles), validation set (60 articles) and test set (60 articles) are identical. The distribution of labels in each set and setting can be found in Table~\ref{table-art-level}. As can be seen, the article-level labels are almost balanced, with some more biased than neutral articles. According to the distribution in the training set, we choose all-bias as the majority baseline in the later experiments.

As standard feature-based approaches, we employ an SVM and a logistic regression classifier based on word $n$-grams with $n \in \{1, 2, 3\}$. The considered $n$-grams are learned on the training set and lowercased. Hyperparameters such as cost and class balance are optimized on the validation set. 

As a standard neural approach, we employ a pre-trained uncased BERT model using word embeddings as ``features''.%
\footnote{Cased and uncased BERT performed similarly in tests.}  
We fine-tuned the approach and optimize the number of epochs for fine-tuning on the training and validation set. Only the first 256 and the last 256 words of an article are used for bias prediction, because the maximum sequence length of the BERT model is 512 tokens.

\paragraph{Q2.} 

To study Q2, we use the same splitting of articles as used for the \emph{w/o event} setting above. In the experiments of this research question, we use the ground-truth sentence-level bias from the dataset. Thereby, we investigate the ideal case where the sentence-level bias can be detected perfectly (assuming the manual annotations are correct). The different types of sentence-level bias are also tested to understand if article-level bias is more correlated to a certain type.

We prepare three types of sentence-level bias features, according to the descriptions in Section~\ref{sec:approach}: 
For {\em bias frequency}, we consider a single feature SVM. We use linear kernel and optimize its cost hyperparameter on the validation set. 
For {\em bias positions}, we compute the bias probability in each position and then apply either Na\"{i}ve Bayes, in line with Equation~\ref{equation:nb}, or an SVM. 
For {\em bias \em sequences}, we use the Markov process from Equation~\ref{equation:markov} to predict an article-level bias label. Besides, we use the probabilities $p(\bar{b}_i\mid \bar{b}_{i-1}, a)$ as features for an SVM. 
Finally, we also test {\em stacking} models. To test the effectiveness of each feature, we stack all three SVMs of each bias feature, as well as any two of the three SVMs as an ablation test.

\paragraph{Q3.} 

To study Q3, we test our approach in a real-world scenario. We first employ the same features and models as in Q1 for sentence-level bias classification. The only difference between article-level and sentence-level setting is that we do not trim sentences for the BERT model. The best classifier is later used in subsequent experiments. The splitting of sentences follows the \emph{w/o event} splitting in the article-level bias detection, i.e., the sentences in the training set represent are used for training, and accordingly for validation and test. The distribution of the different types of sentence-level bias in each set can be found in Table~\ref{table-sen-level}.

\begin{table}[t!]

\small
\centering
\setlength{\tabcolsep}{1.4pt}
\begin{tabular*}{\linewidth}{lrrrrrr}
\toprule
& \multicolumn{2}{c}{\bf Training} & \multicolumn{2}{c}{\bf Validation} & \multicolumn{2}{c}{\bf Test}\\
\cmidrule(l@{2pt}r@{2pt}){2-3} \cmidrule(l@{2pt}r@{2pt}){4-5} \cmidrule(l@{2pt}r@{2pt}){6-7}
& Neutral & Bias & Neutral & Bias & Neutral & Bias\\
\midrule
Lexical bias & 4\,611 & 263 & 1\,558  & 85 & 1\,382 & 78\\
Informational bias & 4\,102 & 772 & 1\,404  & 239 & 1\,272 & 188\\
Any bias & 3\,839 & 1035 & 1\,319  & 324 & 1\,194 & 266\\
\bottomrule
\end{tabular*}
\caption{Distribution of the different types of sentence-level bias in the settings for research question~Q1. In the \emph{Any bias} setting, a sentence is considered biased if it contains lexical and/or informational bias.}
\label{table-sen-level}
\end{table}

Given the predicted sentence-level bias from Q1, we test our approaches as in~Q2. Also, we test a scenario where the event information is available. Similar to the setting in Q1, we randomly split the articles and then split the sentences according to their article-level splitting. We then train the sentence-level bias classifiers and use the best one for our approach.

\section{Results and Discussion}
\label{sec:discussion}

To answer the three research questions of this paper, we report and discuss the results of the experiments described in Section~\ref{sec:exp}.

\subsection{Standard Approaches to Bias Detection}

Tables~\ref{table-rq1-w-event} and~\ref{table-rq1-wo-event} show the results of the experiments for Q1, which address the effectiveness of standard classification approaches in article-level bias detection. With a maximum of 0.55, the accuracy of all classifiers is generally low for a two-class classification task. When event information is available, accuracy improves at least up to 10 percentage points over the baseline, though. When not available, the classifiers seem to learn almost nothing: In the absence of event features, the classifiers are more forced to learn style or structural features. Yet, they turn out not to be able to do so without a proper design of such features. These results suggest that standard approaches are insufficient for article-level bias detection.

\begin{table}[t!]
\small
\centering
\setlength{\tabcolsep}{6pt}
\begin{tabular*}{\linewidth}{lll}
\toprule
\bf Feature & \bf Classifier & \bf Accuracy\\
\midrule
-- & All-bias baseline & 0.45 \phantom{($+$0.00)} \\
$n$-grams (1--3) & SVM & 0.55 ($+$0.10) \\
$n$-grams (1--3) & Logistic Regression & 0.46 ($+$0.01) \\
Word embeddings & BERT & 0.52 ($+$0.07) \\
\bottomrule
\end{tabular*}
\caption{Accuracy of the three standard approaches and the all-bias baseline in article-level bias deteciotn on the dataset split {\em w/ event}. The numbers in parentheses indicate the difference compared to the baseline.}
\label{table-rq1-w-event}
\end{table}

\begin{table}[t!]
\small
\centering
\setlength{\tabcolsep}{6pt}
\begin{tabular*}{\linewidth}{lll}
\toprule
\bf Feature & \bf Classifier & \bf Accuracy\\
\midrule
-- & All-bias baseline & 0.52 \phantom{($+$0.00)} \\
$n$-grams (1--3) & SVM & 0.52 ($+$0.00) \\
$n$-grams (1--3) & Logistic Regression & 0.53 ($+$0.01) \\
Word embeddings & BERT & 0.53 ($+$0.01) \\
\bottomrule
\end{tabular*}
\caption{Accuracy of the three standard approaches and the all-bias baseline in article-level bias detection on the dataset split {\em w/o event}. The numbers in parentheses indicate the difference compared to the baseline.}
\label{table-rq1-wo-event}
\end{table}

\subsection{Impact of Sentence-Level Bias in General}

\begin{table}[t!]

\small
\centering
\setlength{\tabcolsep}{2pt}
\begin{tabular*}{\linewidth}{lllrr}
\toprule
\bf Bias & \bf Feature & \bf Classifier & \bf Acc (GT) & \bf Acc (Pr)\\
\midrule
Lex. & $f_{abs}$ &  SVM & {\bf 0.65} & {\bf 0.52} \\
& $f_{rel}$  & SVM & 0.63 & 0.48\\
\addlinespace[3px]
& Bias Position & Na\"{i}ve Bayes & 0.55 & 0.48 \\
& & SVM & 0.57 & 0.48 \\
\addlinespace[3px]
& Bias Sequence & Markov Process & 0.50 & 0.50 \\
& & SVM & 0.53 & 0.50 \\
\cmidrule{2-5}
& F + P & SVM Stacking & {\bf 0.65} & {\bf 0.52} \\
& F + S & SVM Stacking & {\bf 0.65} & {\bf 0.52} \\
& P + S & SVM Stacking & 0.52 & {\bf 0.52} \\
& F + P + S & SVM Stacking & {\bf 0.65} & {\bf 0.52} \\
\midrule
Info. & $f_{abs}$ & SVM & 0.57 & 0.52 \\
& $f_{rel}$  & SVM & 0.52 & 0.52 \\
\addlinespace[3px]
& Bias Position & Na\"{i}ve Bayes & 0.55 & 0.50\\
& & SVM & 0.55 & 0.50 \\
\addlinespace[3px]
& Bias Sequence & Markov Process & 0.48 & 0.48 \\
& & SVM & 0.47 & 0.48 \\
\cmidrule{2-5}
& F + P & SVM Stacking & 0.55 & 0.52 \\
& F + S & SVM Stacking & {\bf 0.58} & 0.52 \\
& P + S & SVM Stacking & {\bf 0.58} & 0.52 \\
& F + P + S & SVM Stacking & {\bf 0.58} & {\bf 0.57} \\
\midrule
Any & $f_{abs}$ & SVM & 0.65 & {\bf *0.67} \\
& $f_{rel}$  & SVM & 0.65 & 0.65\\
\addlinespace[3px]
& Bias Position & Na\"{i}ve Bayes & 0.57 & 0.58\\
& & SVM & 0.52 & 0.52\\
\addlinespace[3px]
& Bias Sequence & Markov Process & 0.58 & 0.58\\
& & SVM & 0.42 & 0.42\\
\cmidrule{2-5}
& F + P & SVM Stacking & 0.63 & 0.65\\
& F + S & SVM Stacking & {\bf *0.67} & 0.62\\
& P + S & SVM Stacking & 0.50 & 0.50\\
& F + P + S & SVM Stacking & {\bf *0.67} & 0.62\\
\bottomrule
\end{tabular*}
\caption{Accuracy of all evaluated combinations of features and classifiers in article-level bias detection based on ground-truth (GT) and predicted (Pr) sentence-level bias. F combines absolute ($f_{abs}$) and relative ($f_{rel}$) bias frequency, P stands for for bias position, and S for bias sequence. The best value for each bias type is marked bold. The best values overall are marked with~*.}
\label{table-rq23}
\end{table}

As regards~Q2, the column \emph{Acc(GT)} of Table~\ref{table-rq23} shows the accuracy of employing ground-truth sentence-level bias features in predicting article-level bias. The SVM stacking classifier with bias frequency and sequence (F+S) performs best with an accuracy of~0.67. Stacking all features (F+P+S) achieves the same accuracy. In general, all feature and classifier combinations outperform all approaches found in Table~\ref{table-rq1-wo-event}.

Among the features for sentence-level bias, bias frequency and bias position can be exploited best by the SVM. While bias sequence does not perform as well as the others, the stacking classifier using it yields the highest effectiveness. The bias sequence appears to be weakest and sometimes brings negative impact to the performance. However, there may be several reasons behind it. For example, the sequential features may be too subtle, such that our models (SVM and Markov process) are too sensitive to the tiny changes in the features. But, it may also be that a smarter combination strategy for the three different types of feature is required; to keep the models simple, we tested only stacking. On the single features, the results show that an SVM is not always the best choice to utilize the features. In particular, Na\"{i}ve Bayes and Markov process work better when dealing with informational bias and any bias. 

Next, we take a closer look at the stacking part of Table~\ref{table-rq23}, to analyze the feature's effectiveness. While using lexically biased sentences as features, the frequency features contribute more (combinations in stacking with F achieve the best results). On the other hand, while using informationally biased sentences as features, the sequential features are more important. In other words, to detect article bias, it is important to know the number of lexically biased sentences as well as the order of informationally biased sentences. Our interpretation is that, the existence of lexical bias is already a strong clue for presenting bias, whereas informational bias has to be conveyed in a certain order or writing strategy (and thus is more difficult to be captured).

Regarding the two types of sentence-level bias, the best results are observed for {\em any} bias. Using only informational bias leads to the lowest effectiveness. While there is more informational than lexical bias, as shown in Table~\ref{table-sen-level}, the classifiers seem to rely more on lexical bias. The reason could be that lexical bias is easier to capture (by the word usage), while informational bias clues, if any, are subtle. Still, including both types of bias (but not distinguishing them) works best.

\subsection{Impact of Predicted Sentence-Level Bias}

\begin{table}[t!]

\small
\centering
\setlength{\tabcolsep}{3pt}
\begin{tabular*}{\linewidth}{lllrr}
\toprule
\bf Bias & \bf Feature & \bf Classifier & \bf Acc. & \bf Prec.\\
\midrule
Lex. & -- &  All-bias baseline & 0.05 & 0.05\\
& $n$-grams (1--3) & SVM & 0.13 & 0.13\\
& $n$-grams (1--3) & Logistic Regression & 0.07 & 0.05 \\
& Word embeddings & BERT & {\bf0.95} & {\bf0.38} \\
\midrule
Info. & -- & All-bias baseline & 0.13 & 0.13 \\
& $n$-grams (1--3) & SVM & 0.13 & 0.13\\
& $n$-grams (1--3) & Logistic Regression & 0.47 & 0.14\\
& Word embeddings & BERT & {\bf0.86} & {\bf0.37} \\
\midrule
Any & -- & All-bias baseline & 0.18 & 0.18 \\
& $n$-grams (1--3) & SVM & 0.38 & 0.18\\
& $n$-grams (1--3) & Logistic Regression & 0.69 & 0.23 \\
& Word embeddings & BERT & {\bf0.79} & {\bf0.58}\\
\bottomrule
\end{tabular*}
\caption{Accuracy (Acc.) and precision (Prec.) of the three standard approaches and the all-bias baseline in sentence-level bias detection. The highest accuracy and precision values for each bias type are marked bold.}
\label{table-rq1-sen}
\end{table}

Regarding Q3, we first present the results of applying the standard approaches to sentence-level bias detection in Table~\ref{table-rq1-sen}. Besides accuracy, we also show precision, since a high precision boosts the confidence in predicting sentence-level bias. We expect precision to be more important than recall, since we use the predicted bias for computing the article-level bias features. We find that fine-tuned BERT is strongest in effectiveness. Matching intuition, predicting lexical bias seems much easier than predicting informational bias.

\begin{table}[t!]

\small
\centering
\setlength{\tabcolsep}{6pt}
\begin{tabular*}{\linewidth}{llrrr}
\toprule
\bf Bias & \bf Classifier & \bf Precision & \bf Recall & \bf F$_1$\\
\midrule
Lex. & \citet{fan:2019} & 29.13 & 38.57 & 31.49\\
& Reimplementation & 37.50 & 13.64 & 20.00 \\
\midrule
Info. & \citet{fan:2019} & 43.87 & 42.19 & 43.27\\
& Reimplementation & 58.62 & 32.08 & 41.46 \\

\bottomrule
\end{tabular*}
\caption{Classification results of \citet{fan:2019} and our reimplementation. Both use pre-trained BERT, but the exact dataset split of \citet{fan:2019} is unclear.}
\label{table-rq1-sen-reimp}
\end{table}

Since \citet{fan:2019} provide their results of using BERT on sentence-level bias classification, we also used BERT for comparison. To this end, we split the dataset into sets of the same {\em size} as~\citeauthor{fan:2019} (randomly with 6819 training, 758 validation, and 400 test instances). However, the actual distribution of labels is not provided by the authors. As shown in Table~\ref{table-rq1-sen-reimp}, the results of our reimplementation for predicting informational bias is comparable to their results (in terms of F$_1$-score), but it is much worse for predicting lexical bias. Note that lexical bias in the dataset is rather rare (478/7984 $\approx$ 6\%). We thus assume that the difference between our and the original test set caused the difference.

We used the predictions of the best sentence-level bias classifier (i.e., BERT) to compute the bias features. The resulting effectiveness in article-level bias detection can be found in column \emph{Acc(Pr)} of Table~\ref{table-rq23}. Comparing these results to those obtained for Q2, we see a clear drop in the effectiveness, when using only lexical bias or only informational bias. Interestingly, however, the best configuration---with absolute bias frequency ($f_{abs}$) and SVM on any bias---is as good as the best one for Q2. This means that using the predicted bias can sometimes be better than using ground-truth bias. We explain this by the fact that sentence-level bias classifiers are deterministic while human annotators may be not, which can help our approaches to learn more stable patterns in the features.

Overall, our approaches with sentence-level bias information clearly outperform the standard approaches, underlining the impact of our approach. With an accuracy of 0.67, we outperform the standard approaches (0.53) by 14 points and the all-bias baseline (0.52) by 15 points. Regarding the different types of bias, the bias frequency is still the best feature, while the bias position and the bias sequence are weaker. The stacking model is the most effective in general. 

Finally, we also considered the case where event information is available, as in Table~\ref{table-rq1-w-event}. We followed the same process by selecting the best sentence-level bias classifier, which is again BERT with 0.83 accuracy and 0.58 precision, and use it to generate the article-level bias features. Similar to the results in Table~\ref{table-rq23}, the best classifier is an SVM on absolute bias frequency. We achieve 0.60 accuracy outperforming the baseline (0.45), which is again around 15 points higher in accuracy. These results demonstrate that our approach can achieve high effectiveness robustly, regardless of whether it can exploit event information or not.

\subsection{Hyperparameters}

To deepen insights and to simplify reproducibility, this section discusses important hyperparameters used in the experiments.

\paragraph{Bias Normalization}

In the bias position and bias sequence features, the first step is to normalize the length of the bias annotations. Interestingly, the best sampling methods vary in different settings. Specifically, {\em last} is best for bias position with Na\"{i}ve Bayes, {\em average} for bias position with SVM, {\em maximum} for bias sequence with Markov process; and {\em last} for bias sequence with Na\"{i}ve Bayes.

\paragraph{Number of Normalized Positions}

We tested the number of positions needed in the bias position and bias sequence features. This number of positions roughly refers to how many bias clues are in an article. We find that the best value according to the validation set is different in each setting. In summary we determine 10 for bias position with Na\"{i}ve Bayes, 3 for bias position with SVM, 10 for bias position with Markov process, and 8 for bias position with SVM.

\paragraph{Number of Gaussian Mixtures}

The number of Gaussian mixtures indicates the variability of the bias distribution in a single position. We find that the best number of mixtures is 3 for bias position with SVM, and 5 for other settings. While this value depends also on the number of datapoints, it shows that setting it to 5 mixtures is reasonable in general.

\paragraph{Number of Markov's Order}

We tested the order of the Markov process in Equation~\ref{equation:markov}. We find that first-order Markov (a position depends on the previous position only) is best. As discussed, longer dependencies require more datapoints to estimate a better transition probability. Due to the size of our dataset (300 articles with 180 of them as the training set), the second or higher order of Markov does not make sense in our case.

\section{Conclusion}
\label{sec:conclusion}

In this paper we have given evidence that the exploitation of low-level lexical information is insufficient to detect article-level bias --- especially, if the dataset is small. To provide a complete picture, we have formulated three research questions related to article-level bias detection, in order
\Ni
to assess the state of the art of event-dependent and event-independent bias prediction,
\Nii
to learn about the relation between sentence-level and article-level bias, and
\Niii
to study whether sentence-level bias can be leveraged to predict article-level bias.

To tackle the detection of article-level bias, we have proposed and analyzed derived (second-order) bias features, including bias frequency, bias position, and bias sequence. As a main result of our research, we have shown that this new approach clearly outperforms the best approaches existing so far.

If bias detection can be done sufficiently robust on article level, we envisage, as a line of future research, the development of ``reformulation'' strategies and algorithms for the task of neutralizing biased articles \cite{pryzant:2020}.

\section*{Acknowledgments}

This work was partially supported by the German Research Foundation (DFG) within the Collaborative Research Center ``On-The-Fly Computing'' (SFB~901/3) under the project number~160364472.

\begin{raggedright}
\bibliography{emnlp20-bias-detection-lit}
\bibliographystyle{acl_natbib}
\end{raggedright}

\end{document}